\documentclass[runningheads]{llncs}

\usepackage{amsmath,amsfonts,bm}

\def\eqref#1{equation~\ref{#1}}

\def\1{\bm{1}}

\DeclareMathAlphabet{\mathsfit}{\encodingdefault}{\sfdefault}{m}{sl}
\SetMathAlphabet{\mathsfit}{bold}{\encodingdefault}{\sfdefault}{bx}{n}

\usepackage{pgfplots}
\pgfplotsset{height=5.2cm} 
\pgfplotsset{compat=1.7}

\usepackage{times}
\usepackage{soul}
\usepackage{url}
\usepackage[hidelinks]{hyperref}
\hypersetup{
    pdftitle={Scaling up graph homomorphism for classification via sampling},
    bookmarks=true,
    pdfpagemode=FullScreen,
}

\usepackage[utf8]{inputenc}
\usepackage[small]{caption}
\usepackage{amsmath}
\usepackage{amssymb}
\usepackage{booktabs}
\usepackage{algorithm}
\usepackage{algorithmic}
\usepackage{multirow}
\usepackage{comment}
\usepackage{dsfont}

\usepackage{comment}
\usepackage[colorinlistoftodos,prependcaption]{todonotes}

\DeclareMathOperator{\Hom}{Hom}

\DeclareMathOperator{\Prob}{\mathbb{P}}
\DeclareMathOperator{\tw}{tw}

\usepackage{siunitx}
\usepackage{mathtools}

\title{Scaling up graph homomorphism \\ for classification via sampling}

\author{Paul Beaujean \and Florian Sikora \and Florian Yger}
\institute{LAMSADE, Universit\'e Paris-Dauphine, PSL \\
\texttt{\{paul.beaujean,florian.sikora,florian.yger\}@lamsade.dauphine.fr}}

\begin{document}

\maketitle

\begin{abstract}
Feature generation is an open topic of investigation in graph machine learning. In this paper, we study the use of graph homomorphism density features as a scalable alternative to homomorphism numbers which retain similar theoretical properties and ability to take into account inductive bias. For this, we propose a high-performance implementation of a simple sampling algorithm which computes additive approximations of homomorphism densities. In the context of graph machine learning, we demonstrate in experiments that simple linear models trained on sample homomorphism densities can achieve performance comparable to graph neural networks on standard graph classification datasets. Finally, we show in experiments on synthetic data that this algorithm scales to very large graphs when implemented with Bloom filters.
\end{abstract}

\section{Introduction}

Statistical learning was first developed to study the performance of predictive models on Euclidean data where each observation is a point in some vector space. %
In domains where the focus is relationships between entities, data is often represented by graphs of varying size and connectivity. This mismatch between models operating on vectors and graph data makes it difficult to adapt existing machine learning techniques to certain application domains. In particular in chemistry and biology, many structures such as molecules~\cite{gauzere2014graph}, protein interactions, interactions between genes, and even brain regions~\cite{networkBCI} are naturally modeled as graphs. 

\paragraph{Graph embeddings}

There are two main ways to find vector representations for non-Euclidean data: either computing an embedding of the data into a vector space directly, or learning this embedding~\cite{bronstein2017geometric}. For example, the bag of words model is a standard method in natural language processing which maps a sentence to the number of occurrences of each word that appears in it. This usually sparse high-dimensional vector is then amenable to further preprocessing or directly used as an input to machine learning models.
On the other hand, richer embeddings that are obtained by learning such as word2vec~\cite{mikolov2013distributed} or Glove~\cite{pennington2014glove} often outperform direct embeddings and improve the accuracy of several downstream machine learning tasks. However, their learning procedures require significantly more computational resources than what is required to compute simple embeddings like the bag of words model and to train simple models on the resulting vectors.

There is ongoing research directed towards finding invariant vector representations of graphs and studying their properties~\cite{grohe2020word2vec}. Among computed embedding methods, a popular one is to use for each graph the histogram of colors obtained in the canonical coloring given by the Weisfeiler--Leman color refinement algorithm~\cite{weisfeiler1968reduction}. Such histograms of colors serve as vector representations which are guaranteed to correctly detect isomorphic graphs but can fail to tell apart two non-isomorphic graphs.
While color refinement provides global information about the graph, other embedding approaches aim to represent a graph via its substructures. In particular, subgraphs have been a main point of interest in the design of features used in machine learning tasks on graphs. %
Shervashidze et al. have proposed kernel methods based on graphlet statistics \cite{shervashidze2009efficient} both with exact counting algorithms and sampling alternatives which amount to studying the distribution of small induced subgraphs. NT and Maehara have studied the use of graph homomorphism numbers as graph embeddings \cite{ghc-nt20} which can be seen as counting a specified collection of partial subgraphs. In the graph kernel literature, statistical information regarding random walks or specific subgraphs such as shortest paths and subtrees has been the basis of several methods which can be computed efficiently \cite{kriege2020survey}.
In the field of learned graph embeddings, graph neural networks remain the most popular technique with a large part of the literature being focused on finding the best variants of message-passing graph neural networks~\cite{scarselli2008graph}.

In this work, we contribute to the study of feature generation for graph data, i.e. we directly compute invariant graph embeddings that will be given as inputs to standard machine learning operating on Euclidean data~\cite{cai2018comprehensive}. Good embeddings should be fast to compute, have high representational power, and ideally allow the user to factor in inductive bias.

\paragraph{Limitations of homomorphism numbers}
NT and Maehara~\cite{ghc-nt20} leverage results by Lov\'asz \cite{lovasz2012large} on graph homomorphisms to promote the use of homomorphism numbers as graph embeddings that carry desirable theoretical guarantees. In particular they introduce a class of embeddings parameterized by a collection of pattern graphs or motifs $\mathcal{F}$. Counting graph homomorphisms from the patterns $\mathcal{F}$ corresponds to counting partial subgraphs of that collection. 

Graph homomorphism numbers are closely related to several classes of existing graph machine learning techniques. Given an appropriate choice of $\mathcal{F}$ they are equivalent to color refinement algorithms \cite{dell2018lovasz} and to a large class of graph neural networks \cite{morris2019weisfeiler}. Moreover, NT and Maehara~\cite{ghc-nt20} prove universality theorems which state that if a function operating on graphs is invariant over graphs that have the same homomorphism numbers from $\mathcal{F}$, then it can be accurately approximated by a polynomial of homomorphism numbers from $\mathcal{F}$. 

To highlight the practical use cases of these embeddings, NT and Maehara focus on graph homomorphisms that can be counted efficiently. In particular they develop custom counting methods for tree homomorphisms and cycle homomorphisms which are reminiscent of existing kernel methods. They also give an implementation of the algorithm proposed by D\'iaz et al. \cite{diaz2002counting} which is currently the best algorithm for counting homomorphisms from any arbitrary motif $F$. In particular this algorithm is polynomial if $F$ has bounded treewidth, that is $F$ is close to being a tree.

However, under commonly used computational complexity assumptions, it is not possible to design an algorithm with better worst-case time complexity than the one given by D\'iaz et al.
In practice, this algorithm is fast but cannot be used realistically for large graphs such as those found in large-scale social network datasets \cite{snapnets}.

\paragraph{Our contributions}
To remedy the problem of computing homomorphism numbers on large graphs we attempt to find an alternative that would retain its properties and remain fast to compute at scale.

An homomorphism density is a normalized version of an homomorphism number which, instead of the count, represents the frequency at which a given subgraph $F$ with $k$ nodes appears in a graph $G$ with $n$ nodes. Unlike homomorphism numbers which can grow as large as $n^k$, homomorphism densities are always contained in the (0,1) interval. For this reason, while approximating the number of homomorphisms remains a difficult problem from the point of view of computational complexity, it is still possible to give an additive approximation of homomorphism densities in polynomial time. Note that this only goes one way: an approximate number would give an approximate density but an additive approximation of the homomorphism density does not allow to recover a good approximation of the number of homomorphisms. 

The main contributions of this paper can be summarized as follows.
(1) We demonstrate in graph classification benchmarks on standard datasets that sample homomorphism densities retain the same representational power as homomorphism numbers and simple models using those as features are comparable in performance to popular graph neural network architectures. (2) We show that a high-performance implementation of a simple sampling algorithm which computes an $\varepsilon$-additive approximation of homomorphism densities scales to very large graphs in synthetic experiments. 

\section{An alternative to homomorphism numbers}

In this section we discuss existing theoretical results on graph homomorphisms that justify the relevance of approximate homomorphism densities as graph embeddings.

\subsection{The representational power of homomorphism numbers}
\label{section:homomorphism-numbers-densities}

In the context of graph theory, graph morphisms are functions which map the node set $V(F)$ to $V(G)$. In particular, graph homomorphisms are morphisms that preserve adjacency, e.g. $\forall\, uv \in E(F), f(u)f(v) \in E(G)$. 
A graph homomorphism that also preserves non-adjacency is called a graph isomorphism. We write $G_1 \cong G_2$ when there exists a graph isomorphism between $G_1$ and $G_2$.

The most important fact about graph homomorphisms is that any undirected graph $G$ can be uniquely determined (up to isomorphism) by the set of homomorphisms from pattern graphs to $G$. Following NT and Maehara \cite{ghc-nt20}, we use the notation:
\begin{equation*}
    \Hom(F, G) = \left\{ f : V(F) \to V(G) \mid \forall\, uv \in E(F), f(u)f(v) \in E(G)\right\}
\end{equation*}
for the set of homomorphisms from a pattern graph $F$ to our target graph $G$. The cardinality of this set, $\hom(F,G) = \left|\Hom(F,G)\right|$, is called the homomorphism number of $G$ from $F$. As it turns out, to identify a graph $G$, it is enough to know its homomorphism numbers. 
Since we frequently consider several homomorphism numbers at once it is convenient to gather them in a vector of homomorphism numbers as follows: $\hom_{\mathcal{F}}(G) = (\hom(F,G))_{F \in \mathcal{F}}$ where $\mathcal{F}$ is a set of pattern graphs. This notation allows us to restate the property more formally as follows:
\begin{theorem}{(\cite{lovasz1967operations})}
Given two undirected graphs $G_1$ and $G_2$ with at most $n$ nodes. Denoting by $\mathcal{G}_{n}$ the set of all simple graphs with at most $n$ nodes, we have:
\begin{equation*}
G_1 \cong G_2 \iff \hom_{\mathcal{G}_{n}}(G_1) = \hom_{\mathcal{G}_{n}}(G_2).
\end{equation*}
\label{thm:iso-equiv-same-hom-vector}
\end{theorem}
\vspace{-16pt}
In particular, this theorem can be interpreted as saying that $\hom_\mathcal{G}(G)$ is a canonical vector representation of $G$ that is invariant to labeling, where $\mathcal{G}$ is the set of all simple graphs, i.e. graphs without loops or multiple edges.
It may appear that we have traded a finite graph $G$ for a vector of infinite length $\hom_\mathcal{G}(G)$. However Theorem~\ref{thm:iso-equiv-same-hom-vector} already indicates that to tell apart two graphs with $n$ nodes it is enough to look only at the coordinates of the vector that correspond to pattern graphs with at most $n$ nodes.

A natural question to ask is whether it is possible in practice to consider significantly less than $\sum_{k=1}^n 2^{k(k-1)/2}$ coordinates to separate graphs. In the context of machine learning, this would mean that a short vector of homomorphism numbers could suffice, e.g. for graph classification. NT and Maehara \cite{ghc-nt20} provide a preliminary answer to this question and demonstrate that this approach is adequate even with a small number of small pattern graphs or a few well-chosen patterns graphs.

\subsection{The computational complexity of counting graph homomorphisms}
\label{subsection:hardness-hom}

Even for a single pattern $F$ computing the value $\hom(F,G)$ which is the number of homomorphisms from $F$ to $G$ is $\#P$-hard in general \cite{diaz2002counting} where $\#P$-hard is the analogue of $NP$-hardness for counting problems. This computational complexity is inherited from the hardness of deciding the existence of a specific subgraph in a graph, e.g. finding a large clique. In simpler terms, finding the difference between 0 homomorphism from some $F$ to $G$ and more than 1 homomorphisms is already hard for some $F$. 

For many pattern graphs however, it is still possible to compute homomorphism numbers in polynomial time. For example, if we consider pattern graphs with no more than $k$ nodes, we can obtain a naive polynomial-time algorithm by listing out the $O(n^k)$ subsets of $k$ nodes in $G$ and checking which of these subsets correspond to $F$. This simple approach is often combined with clever heuristics in the graphlet literature to obtain high-performance software to compute subgraph statistics \cite{jamshidi2020peregrine}. 
For specific families of pattern graphs like cycles or trees, there are known polynomial-time algorithms that are significantly more efficient. However, in the case of an arbitrary pattern $F$ the best option is to use the algorithm of D\'iaz et al. \cite{diaz2002counting}. This algorithm relies on a tree decomposition of the pattern graph $F$ and has a worst-case time complexity of $O(\tw(F) \cdot k n^{\tw(F) + 1})$ %
where $G$ and $F$ have respectively $n$ and $k$ nodes and the treewidth $\tw(F)$ is the size of the smallest tree decomposition of $F$ \cite{robertson1986graph}.
This graph parameter, which is $NP$-hard to compute, measures how ``tree-like'' a graph is: lower values signify being very close to trees (which have a treewidth of 1) such as cycles which have treewidth 2 while at the extreme opposite the complete graph $K_n$ has a treewidth of $n - 1$. Unfortunately, this means that even computing the homomorphism number from a relatively small pattern graph e.g. $\hom(K_5, G)$ with the algorithm of D\'iaz et al. is almost as impractical as listing out every subset of 5 nodes.

The algorithm of D\'iaz et al. means that computing homomorphism numbers is tractable when the treewidth parameter is bounded. A natural question is to ask whether there are other cases or parameters that make the problem easier. It turns out that not only this is the only parameter that can make the problem easier \cite{grohe2007} but also that even computing a multiplicative approximation of homomorphism numbers cannot be done efficiently unless the treewidth of $F$ is bounded \cite{bulatov2020}. On a side note, as the number of homomorphisms can range from 0 to $n^k$ it is difficult to define what would be an efficient additive approximation of homomorphism numbers.

This leaves us with two remaining possibilities for computing homomorphism numbers: either we limit ourselves to patterns with low treewidth or we rely on graphs for which a given heuristic would perform well in practice. However there is one blind spot of computational complexity that we can exploit if we switch from homomorphism numbers to homomorphism densities.

\subsection{Efficient sampling for additive approximation of homomorphism density}

Homomorphism densities are normalized homomorphism numbers. Formally, if we write $n = |V(G)|$ and $k = |V(F)|$, we can define the homomorphism density from $F$ to $G$ as follows:
\begin{equation}
    t(F,G) = \hom(F,G) / n^k.
    \label{eq:hom-density}
\end{equation} 
While $\hom(F,G)$ is a magnitude, that is the number of graph homomorphisms from $F$ to $G$, $t(F,G)$ is a proportion. Alternatively, the homomorphism density from $F$ to $G$ is the probability that a morphism $f : V(F) \to V(G)$ drawn uniformly at random is a graph homomorphism, i.e. $t(F,G) = \Prob_{f \sim V(F) \to V(G)} \{f \in \Hom(F,G) \}$. Another way to interpret it is the probability that a random morphism will preserve all edges of the pattern graph $F$ into the target graph $G$.

Like in Theorem~\ref{thm:iso-equiv-same-hom-vector} where we use homomorphism numbers to distinguish between two graphs, we can use homomorphism densities $t_{\mathcal{F}}$ as a canonical graph embedding:
\begin{theorem}{(Isomorphism via homomorphism densities)}
Given two undirected graphs $G_1$ and $G_2$ with the same number of nodes $n$, and $\mathcal{G}_{n}$ the set of all simple graphs with at most $n$ nodes, we have:
\begin{equation*}
G_1 \cong G_2 \iff t_{\mathcal{G}_{n}}(G_1) = t_{\mathcal{G}_{n}}(G_2).
\end{equation*}
\label{thm:iso-equiv-same-t-vector}
\end{theorem}
\vspace{-12pt}
The above can be readily derived from Theorem~\ref{thm:iso-equiv-same-hom-vector} by dividing each equality of homomorphism numbers  $\hom(F_i,G_1) = \hom(F_i,G_2)$ on both sides by $n^{|V(F_i)|}$ for every simple graph $F_i \in \mathcal{G}_n$. %

From section~\ref{subsection:hardness-hom} it is clear that we should not expect homomorphism densities to be easier to compute exactly as multiplying by $n^k$ would suffice to recover the corresponding homomorphism densities. Similarly, if we could obtain a multiplicative approximation of $t(F,G)$ we would obtain the same multiplicative approximation of $\hom(F,G)$ but that is ruled out by hardness results. 

However, since $t(F,G)$ is by definition the probability of a polynomial-time testable property, it is possible to turn sampling into a polynomial-time 
$\varepsilon$-additive approximation algorithm. Such result would be allowed despite the previously mentioned hardness results because there is no guarantee that an additive approximation of $t(F,G)$ can be processed into any meaningful approximation of $\hom(F,G)$. In particular, unless the precision is small enough $\varepsilon < n^k$ there is no way to distinguish between 0 or 1 homomorphism. Furthermore such additive precision would lead to an exponential running time. In other terms, an efficient additive approximation algorithm for homomorphism densities would not be guaranteed to detect very small densities but would produce accurate estimates for larger densities.

\section{Implementing a high-performance sampling algorithm for approximate homomorphism density}

In this section we describe an efficient parallel implementation of a sampling algorithm which gives a $\varepsilon$-additive approximation of homomorphism densities and we give an analysis of its time complexity. This algorithm relies on standard sampling arguments similar to those used in sketching algorithms for graph data~\cite{ahn2012graph}.

As earlier, we say that $G$ and $F$ have respectively $n$ and $k$ nodes. Notice that we can sample a morphism $f : V(F) \to V(G)$ uniformly at random from the space of all functions $V(F) \to V(G)$ by drawing $k$ integers independently at random from the uniform discrete distribution $\mathcal{U}(0, n-1)$. Here we represent $f$ by an array of size $k$ where function application $f(u)$ is realized by accessing the array $f[u]$. Furthermore, sampling $N$ such morphisms can be done by requesting a $N$-by-$k$ array instead. Sampling uniform integers is implemented in every high-performance pseudo-random number generator software suite such as the \texttt{randint} routine provided by the NumPy library \cite{harris2020array}. We are left with computing the sample mean for the probability of $f$ to be a homomorphism.
  
However, for the sample mean to be close to the actual mean, we need to make sure to compute it from a large enough number of samples.
In essence, estimating an homomorphism density $t(F,G)$ is identical to estimating the unknown bias of a coin, i.e. a Bernoulli distribution of parameter $p \in [0,1]$. Here, a random morphism corresponds to a coin flip which lands on heads when the morphism is an homomorphism, while the density $t(F,G)$ corresponds to the unknown parameter $p$ of the biased coin.

Fortunately, it is well known that we can to use Chernoff bounds \cite{chernoff1952measure} to derive a number of samples that is sufficient to reach an $\varepsilon$-additive approximation of $p$. This leads to the following folklore result:
\begin{theorem}{(Sampling lemma)}
Let $X_1, \ldots, X_N$ be $N$ i.i.d samples of a Bernoulli distribution with parameter $p \in [0,1]$. The following implication holds:
\begin{equation*}
    N \geq \dfrac{1}{2\varepsilon^2} \log  \dfrac{2}{\delta}
    \implies
    \Prob\left(
    \left|\frac{1}{N}\sum_{i=1}^N X_i - p\right| > \varepsilon
    \right) \leq \delta
\end{equation*}
where $\varepsilon > 0$ is the additive precision of our sampling and $1 - \delta \in (0,1)$ the degree of confidence in our estimate.
\label{thm:sampling-lemma}
\end{theorem}

\begin{algorithm}
\begin{algorithmic}[1]
\REQUIRE $G$ an undirected graph on $n$ nodes, $F$ a pattern graph on $k$ nodes and $l$ edges, $\varepsilon > 0$ the requested additive precision, $1 - \delta \in (0,1)$ the desired confidence.
\ENSURE $\bar{t}$ such that $\Prob(|t(F,G) - \bar{t}| > \varepsilon) \leq \delta$
\STATE $N \gets O(\varepsilon^{-2} \log \delta^{-1})$
\FOR{ $i = 1$ to $N$ } \label{alg:rng}
\STATE $f_i \sim \left(\mathcal{U}(0, n-1)\right)_{[k]}$
\ENDFOR
\STATE $\bar{t} \gets \frac{1}{N} $$\sum_{i=1}^N \prod_{uv \in E(F)} \mathds{1}_{E(G)}(f_i(u)f_i(v)) $ \label{alg:tests}
\RETURN $\bar{t}$
\end{algorithmic}
\caption{Sample homomorphism density}
\label{alg:sample-hom-density}
\end{algorithm}

Let us take a quick look at the worst-case time complexity of Algorithm~\ref{alg:sample-hom-density}. The first step (line~\ref{alg:rng}) amounts to drawing $N$ times $k$ random integers in $\{0,\ldots,n-1\}$ which incurs a running time of $O(N k \log n)$. 
Note that this procedure is embarrassingly parallel and can be split into at least $N$ independent threads.
The second step (line~\ref{alg:tests}) is the bulk of the computational content: for each morphism $f_i$ the algorithm must query $l$ edges of $G$ through an adequate data structure. As soon as an edge query fails, i.e. the morphism fails to preserve that particular edge, the indicator function returns 0 and short-circuits the product. The worst case happens on actual homomorphisms which must query all $l$ edges. The running time of this step is then $O(N l)$ which amounts to a total time complexity of $O\left((k \log n + l)\cdot \varepsilon^{-2} \log\delta^{-1}\right)$. 
While this analysis appears to price the random number generation (line~\ref{alg:rng}) higher than the homomorphism tests (line~\ref{alg:tests}), in practice generators benefits from advanced implementations that can leverage current hardware architectures efficiently. For this reason we focus on optimizing edge queries with adequate data structures described later in this section.

We summarize some properties of our sample homomorphism density algorithm. (1) It can be computed with close to no regard to the size of $G$ as the dependency is in $O(\log n)$. Furthermore, $G$ does not need to fit in memory as we only need: (a) to access its number of nodes to initialize our random number generator and (b) to be able to query random edges. (2) The running time depends linearly on the number of edges in the pattern graph $l = |E(F)|$ which invites us to consider small pattern graphs. (3) Higher precision (smaller $\varepsilon$) is expensive and prohibits the use of the sample homomorphism density to reliably decide the existence of large cliques. We stress again that additive approximation algorithms generally cannot provide useful answers to NP-complete decision problems. (4) A positive sample homomorphism density $\bar{t} > 0$ implies that the real homomorphism density is positive $t(F,G) > 0$, but no quantitative comparison can be given without knowledge of $\varepsilon$ and $\delta$. (5) Higher degrees of confidence are extremely cheap, making it possible to add extra nines to a $\delta = 99.9\%$ without jeopardizing the running time. (6) The free choice of $F$ allows us to emulate existing techniques such as statistics on paths or walks~\cite{kriege2020survey} but also to consider more complex substructures.

\paragraph{Data structures for approximate membership}

The problem of set membership is one of the fundamental building blocks in the study of data structures. A wealth of data structures and algorithms provide various trade-offs and guarantees going from sequential search in an unsorted array, to tries, and hash tables. Of particular note is perfect hashing which allows for constant-time set membership queries at the cost of linear space (the space required to represent a perfect hashing function is proportional to the size of the set). In some way, this space complexity is optimal and cannot be improved upon.

Surprisingly, a simple probabilistic data structure, the Bloom filter~\cite{bloom1970space}, allows to implement set membership close to this optimal space complexity. Each element of the set is represented by no more than a dozen of bits and the membership query is realized in constant time by applying a few hash functions to the element to be queried. Bloom filters are implemented via a array of bits which are filled at indices corresponding to the outputs of the different hash functions. The randomness of this data structure is solely contained in the choice of hash functions and is perfectly deterministic once the hash functions have been chosen.

Bloom filters alleviate the need to store the elements themselves unlike e.g. hash tables and other traditional data structures. This convenience in the specific context of the set membership problem comes at the cost of a false positive rate. Indeed, when a Bloom filter returns that an element is not in the set, that answer is always truthful. However a positive answer does not guarantee that the element is actually in the set.

With regards to computing the sample homomorphism density, each graph $G$ can thus be compressed into a Bloom filter containing information about its edges. In practice this means that each edge, usually represented by a pair of unsigned integers each coded over 32, 64, or 128 bits is compressed down to a constant dozen of bits, regardless of the number of nodes of $G$. Furthermore, each homomorphism test corresponds to computing grouped queries which must all return true. Since the queries are independent, this implies a lower false positive rate for the grouped query than its constituent queries. Finally, the Bloom filter associated with $G$ is constructed once as a read-only data structure which can be queried repeatedly in parallel when computing the sample homomorphism density of $G$ with regards to queries representing the edge set of some pattern graph $F$.

\section{Sample homomorphism densities in graph classification tasks}

We conduct numerical experiments to validate the relevance of sample homomorphism densities as computed graph embeddings for graph classification. We have implemented Algorithm~\ref{alg:sample-hom-density} and our experimental methodology in the Python~3 language leveraging the NetworkX library for graphs \cite{networkx}, the NumPy library for array processing \cite{harris2020array}, and the scikit-learn library for machine learning models \cite{scikit-learn}. We have split our code into two parts, a Python library on one hand and a set of Python scripts that depend on that library on the other. Source code for both parts is provided as a supplementary document.
All our experiments are run on a Linux computer with an 8-core Intel Xeon Skylake processor clocked at 2.2 GHz with hyper-threading and 14 GB of memory. We have not used a GPU for computing embeddings or training models.

We start by describing the settings of our experiments as well as our choice of parameters and datasets. Then, we discuss experimental results that demonstrate the relevance of sample homomorphism densities in the context of graph classification.

\subsection{Sampling in practice}

The guarantees provided by the Chernoff bounds implicitly require the use of an additive precision of $\varepsilon = O(n^{-k})$ to detect the presence of a single partial subgraph, i.e. in the case where $t(F,G) = 1/n^k$. However,  we instead sample at fixed levels of precision, with low values of $\varepsilon \in \{ 0.1, 0.05, 0.01 \}$. 
This choice introduces two biases. 
First, a low precision can be seen as a preprocessing method which filters out low-frequency patterns from the feature vectors, leaving subgraphs that occur at a high frequency in the target graph, e.g. $O(10^{-1})$ or $O(10^{-2})$, and for which we can guarantee a good additive approximation. 
Second, fixed precision implies low precision on larger pattern graphs and higher precision on smaller ones which in some extreme cases can lead to sampling being equivalent to exhaustive search, e.g. the target graph is relatively small and the pattern graph has 2 or 3 nodes. 
This means that only highly frequent larger pattern graphs can appear in the feature vectors while smaller pattern graphs are detected even at lower frequencies. 

As mentioned above, the cost of achieving high confidence is extremely low compared to that of precision. However, exploratory testing reveals that confidence has virtually no impact on graph classification tasks so we keep it fixed at $\delta = 95\%$.

\subsection{Pattern graphs and weighted homomorphisms}

NT and Maehara \cite{ghc-nt20} focus their study on homomorphism numbers for specific families of patterns, such as trees and cycles. This choice is partly motivated by the lower computational complexity of computing homomorphism numbers for these families with the $O(\tw(F) \cdot k n^{\tw(F) + 1})$ algorithm of D\'iaz et al. \cite{diaz2002counting}, as trees and cycles have respectively a treewidth of 1 and 2.
On the other hand, since Algorithm~\ref{alg:sample-hom-density} does not depend on low treewidth or on a modest size of $G$, we choose to consider graph patterns among small connected graphs of the Atlas of Graphs provided by the NetworkX library \cite{networkx}. The first 10 graphs of this list correspond to all connected undirected graphs with at most 4 nodes. The next 10 graphs are connected undirected graphs with 5 nodes (out of 21 such graphs) and contain several graphs of treewidth 3.
Each component of the feature vectors thus corresponds to a graph in that list. 
To study the impact of information from larger subgraphs we vary our family of pattern graphs $\mathcal{F}$ with increasing numbers of subgraphs with the following values: 10, 15, 20. 

We consider three variants of homomorphism densities, the original unweighted variant, a weighted variant which assigns to each morphism the product of the numerical attributes of the target nodes, and finally a second weighted variant which disregards node attributes if they exist and instead weighs each node by its degree.

\subsection{Model evaluation methodology}

We follow standard model evaluation methodology with a 10-fold cross-validation procedure closely following the study of Errica et al. \cite{errica2019fair}. Each dataset is first split in 10 blocks, with an inner holdout procedure splitting each block with a 4:1 training/validation ratio. Each block is validated independently and once a set of hyper-parameters has been selected, the corresponding model is trained and scored 3 times to smooth out the randomness that could be caused by different initial conditions during the training procedure.

However, since our underlying features are the result of a sampling procedure, we also average the entire 10-fold cross-validation procedure over 10 independent samples obtained for each dataset. Reported cross-validation scores, e.g. mean over 10 blocks of the test accuracy of the best models, are reported as averages over these 10 independent samples, and the standard deviation reported is the mean over 10 samples of their corresponding standard deviations.

To perform sensitivity analysis over the different parameters of the sampling procedure mentioned above, we hold every parameter constant except the topic of interest. 
All accessible random seeds are accounted for and controlled, both in the model evaluation procedure as well as in the feature generation code. 
Nevertheless, subroutines present in the scikit-learn library \cite{scikit-learn} carry randomness that we cannot control which leads to an unavoidable variance in our experiment results. 
From exploratory testing, this variance is only visible in the third decimal of our reported scores which is why we decide to not perform an in-depth analysis of its effect. 

\subsection{Models and hyperparameters}

The features we consider simply concatenate the number of nodes of a given graph with an approximate homomorphism density vector. These features are then given as inputs to multiple machine learning models for classification. Because we study the properties of homomorphism density features and are not attempting at challenging the state of the art in graph classification, we focus on simple models that are well understood.
The simplest model we consider is a logistic regression classifier with $\ell_1$, $\ell_2$, or no regularization. We select the regularization parameter $C$ among the values: $10^{-4}, 10^{-2}, 10, 10^{4}$. Training is done using the \texttt{liblinear} solver through the interface provided by scikit-learn. We also study several non-linear models %
which we describe in further details in the supplementary material.

\subsection{Datasets}

To compare our method with existing research, we use publicly available datasets from the TUDataset collection \cite{Morris+2020} which can be split in two groups. First, we consider datasets of graphs obtained from biochemistry applications such as MUTAG \cite{debnath1991mutag}. Note that the MUTAG dataset, widely used in the graph machine learning literature, only contains 188 graphs with 17.9 nodes on average.
This small dataset size artificially inflates the variance of cross-validation procedures (a single error costs 5.3\% test accuracy). We also consider datasets of with a larger number of graphs such as NCI1~\cite{wale2008nci1} (4110 graphs, 29.9 nodes on average, 0.2\% per error), ENZYMES (600 graphs, 36.2 nodes on average, 1.7\% per error) and PROTEINS~\cite{borgwardt2005proteins} (1113 graphs, 39.1 nodes on average, 0.9\% per error), or DD~\cite{dobson2003dd} (1178 graphs, 284.3 nodes on average, 0.8\% per error). 
A second group consists of datasets obtained from online social networks, with the COLLAB (5000 graphs, 74.5 nodes on average, 0.2\% per error), REDDIT-BINARY (2000 graphs, 429.6 nodes on average, 0.5\% per error), and IMDB-BINARY datasets (1000 graphs, 19.7 nodes on average, 1.0\% per error)~\cite{yanardag2015social-network-datasets}.

\begin{table*}[t!]
\centering
\resizebox{0.99\linewidth}{!}{%
{ %
\renewcommand{\arraystretch}{1.15}
\begin{tabular}{l|ccccc|ccc}
              & \textbf{MUTAG}        & \textbf{NCI1}         & \textbf{PROTEINS}     & \textbf{DD}           & \textbf{ENZYMES}      & \textbf{REDDIT-B}     & \textbf{COLLAB}       & \textbf{IMDB-B}      \\
\hline
SGHD-$\mathcal{A}_{10}$ $\varepsilon=0.1$  & $83.6\pm8.7$ & $62.6\pm2.9$ & $72.0\pm4.1$ & $76.2\pm3.2$ & $21.2\pm4.1$ & $73.6\pm3.2$ & $68.1\pm2.0$ & $63.3\pm3.7$  \\ %
SGHD-$\mathcal{A}_{10}$ $\varepsilon=0.01$  & $86.3\pm7.9$ & $62.7\pm3.1$ & $72.3\pm3.6$ & $76.1\pm3.2$ & $26.3\pm4.5$ & $73.8\pm3.2$ & $67.7\pm2.1$ & $68.7\pm2.4$ \\ %
\hline
GHC-$\mathcal{T}_{13}$ (NT\&M.) & $88.2\pm7.4$ & $65.4\pm2.5$ & $70.6\pm4.7$ & $75.3\pm3.6$ & $21.7\pm3.6$ & $84.6\pm2.3^*$ & $62.1\pm1.9$ & $69.7\pm4.4$  \\ %
GHC-$\mathcal{C}_{7}$ (NT\&M.)  & -            & -            & -            & $76.1\pm3.9$ & $29.5\pm3.2$ & -            & -            & -           \\ %
\hline
\hline
DGCNN  (Errica+)       & -            & $76.4\pm1.7$ & $72.9\pm3.5$ & $76.6\pm4.3$ & $38.9\pm5.7$ & $77.1\pm2.9$ & $57.4\pm1.9$ & $53.3\pm5.0$  \\ %
DiffPool (Errica+)     & -            & $76.9\pm1.9$ & $73.7\pm3.5$ & $75.0\pm3.5$ & $59.5\pm5.6$ & $76.6\pm2.4$ & $67.7\pm1.9$ & $68.3\pm6.1$  \\ %
ECC   (Errica+)        & -            & $76.2\pm1.4$ & $72.3\pm3.4$ & $72.6\pm4.1$ & $29.5\pm8.2$ & OOR          & OOR          & $67.8\pm4.8$  \\ %
GIN (Errica+)      & -            & $80.0\pm1.4$ & $73.3\pm4.0$ & $75.3\pm2.9$ & $59.6\pm4.5$ & $\textbf{87.0}\pm4.4$ & $\textbf{75.9}\pm1.9$ & $66.8\pm3.9$ \\ %
GraphSAGE (Errica+)    & -            & $76.0\pm1.8$ & $73.0\pm4.5$ & $72.9\pm2.0$ & $58.2\pm6.0$ & $86.1\pm2.0$ & $71.6\pm1.5$ & $69.9\pm4.6$  \\ %
\hline
No Topology (Errica+)  & -            & $69.8\pm2.2$ & $\textbf{75.8}\pm3.7$ & $\textbf{78.4}\pm4.5$ & $\textbf{65.2}\pm6.4$ & $72.1\pm7.8$ & $55.0\pm1.9$ & $50.7\pm2.4$ \\ %
\hline
GIN (NT\&M.)      & $75.5\pm7.6$ & $76.7\pm1.2$ & $73.0\pm1.8$ & -          & -          & $74.1\pm2.3$ & $\textbf{75.9}\pm0.8$ & $70.7\pm1.1$ \\ %
GNTK (NT\&M.)     & $\textbf{89.5}\pm7.1$ & $\textbf{83.8}\pm1.6$ & -          & -          & -         & -          & -          & $\textbf{75.6}\pm3.9$ \\ %
\end{tabular}
}
}
\caption{Test accuracy scores. OOR: out of resources during training, ``-'': experiment not provided in the corresponding study.}
\label{tab:us-vs-errica-vs-nt}
\end{table*}

\subsection{Results and discussion}
We compare the test accuracy of simple linear models trained on sample homomorphism density features with recent studies in graph classification realized with a sound methodology. Numerical results are summarized in Table~\ref{tab:us-vs-errica-vs-nt}.

The first category of models corresponds to those trained on embeddings computed from the graph topology without taking into account node features. Among the numerous variants of our proposed method we report scores of the simplest model: SGHD-$\mathcal{A}_{10}$. The Sample Graph Homomorphism Density $\mathcal{A}_{10}$ model is a logistic regression classifier trained on sample homomorphism densities computed from 10 motifs of the Atlas of graphs with coarse or medium fixed precision. Note that features are given as inputs to the model without preprocessing. Directly related to our method is the study of homomorphism numbers for graph classification of NT and Maehara~\cite{ghc-nt20} which follows a similar model evaluation methodology except for the lack of repeated scoring of selected models. We consider two of their SVC models based on their Graph Homomorphism Convolution framework: GHC-$\mathcal{T}_{13}$ which computes homomorphism numbers from the first 13 trees, and GHC-$\mathcal{C}_{7}$ from the first 7 cycles.

On the other hand, models in the second category learn embeddings from node features with implicit access to graph topology. Since our model evaluation methodology follows closely the work of Errica et al.~\cite{errica2019fair} we have high confidence in comparing our test accuracy scores with the ones they report. The study of Errica et al. gives a fair and reproducible assessment of the performance of popular graph neural network architectures (DGCNN, DiffPool, ECC, GIN, GraphSAGE) which aggregate node features via a message-passing training procedure in comparison to a baseline (``No Topology'') that disregards the graph topology information and only aggregates node features. Furthermore we also include a different implementation of GIN together with the Graph Neural Tangent Kernel method as tested in~\cite{ghc-nt20}.

On most datasets we observe that our SGHD-$\mathcal{A}_{10}$ models achieve similar or better performance than the GHC models based on homomorphism numbers. Furthermore, test accuracy scores are very close to those of multiple GNN architectures which were trained over a maximum of 2 days.

However, it is clear that the dataset NCI1 cannot be properly learned with graph homomorphism techniques, and that sample homomorphism densities lose some information compared to homomorphism numbers. Similarly, the ENZYMES dataset with 6 classes cannot be learned from the graph homomorphism features utilized here although increasing accuracy or specializing the patterns towards cycles does improve performance slightly. We note that two GNN architectures also fail to classify the ENZYMES dataset.

We notice that our SGHD models lose to GHC models on 2 datasets, with an 11 points difference on IMDB-B. It is interesting to note that in the supplementary material to~\cite{ghc-nt20} we find that GHC-$\mathcal{T}_{13}$ achieves a very similar $73.8\% \pm 2.8\%$ test accuracy when trained after min/max or max/abs preprocessing while it achieves over 80\% accuracy with other scaling techniques (standard, quantile, power). This hints at the possibility that additional preprocessing or models with higher capacity may be able to extract more information from sample homomorphism densities.

The following observations are taken from experimental results detailed in our supplementary material. We observe a slight increase in test accuracy from considering weighted homomorphisms for SGHD on biochemistry datasets in every combination of parameters. On social datasets, which do not have node features, there is a strong positive effect of sampling at higher precision. This effect is compounded if we consider higher precision together with more graph patterns.

Finally, we acknowledge that while comparable to some learned embeddings, our models using computed embeddings are less competitive than the best methods that rely primarily, or even exclusively on node features. This latter fact constitutes a challenge to the field of graph machine learning.

\section{Testing the scalability of sample homomorphism densities}

As mentioned previously, standard benchmark datasets in graph classification such as those featured in the TUDataset collection \cite{Morris+2020} as well as more recent large-scale datasets such as the Open Graph Benchmark \cite{hu2020open} or the datasets introduced by Dwivedi et al. \cite{dwivedi2020benchmark} only contain graphs which have at most a few hundred nodes. While these modern datasets include significantly more graphs per dataset, each graph remains small. For example, the recent OGB Large Scale Challenge \cite{hu2021ogblsc} introduces a new dataset for graph classification which is made of 3.8M graphs each containing on average 14.5 edges. Our method on the other hand is designed to be used on large graphs and because of the lack of corresponding dataset, we settle for experiments conducted on synthetic graphs. 

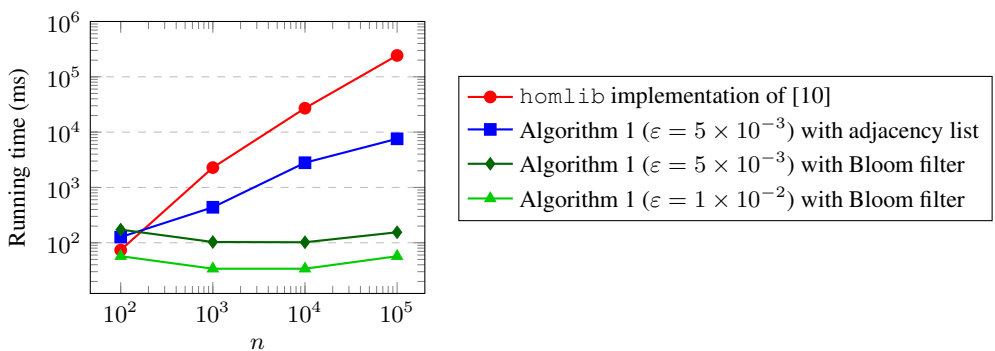
\begin{figure}
\begin{tikzpicture}[trim right=0cm]
\begin{loglogaxis}[
    xlabel={$n$},
    ylabel={Running time (ms)},
    xmax=200000,
    ymax=1000000,
    xtick={100,1000,10000,100000},
    ytick={10,100,1000,10000,100000,1000000},
    legend style={at={(1.1,0.8)},anchor=north west},
    every axis plot/.append style={thick},
    legend cell align={left},
    ymajorgrids=true,
    grid style=dashed,
]

\addplot[
    color=red,
    mark=*,
    ]
    coordinates {
    (100,73.8)
    (1000,2285)
    (10000,27062)
    (100000,243190)
    };
    \addlegendentry{\texttt{homlib} implementation of~\cite{diaz2002counting} }

\addplot[
    color=blue,
    mark=square*,
    ]
    coordinates {
    (100,126.5)
    (1000,438.1)
    (10000,2795.6)
    (100000,7572.3)
    };
    \addlegendentry{Algorithm~\ref{alg:sample-hom-density} ($\varepsilon = \num{5e-3}$) with adjacency list}

\addplot[
    color=green!40!black,
    mark=diamond*,
    ]
    coordinates {
    (100,172.8)
    (1000,103.1)
    (10000,102.1)
    (100000,154.7)
    };
    \addlegendentry{Algorithm~\ref{alg:sample-hom-density} ($\varepsilon = \num{5e-3}$) with Bloom filter}

\addplot[
    color=green!80!black,
    mark=triangle*,
    ]
    coordinates {
    (100,57.2)
    (1000,33.93)
    (10000,33.93)
    (100000,57.08)
    };
    \addlegendentry{Algorithm~\ref{alg:sample-hom-density} ($\varepsilon = \num{1e-2}$) with Bloom filter}
\end{loglogaxis}
\end{tikzpicture}
\caption{Running time of homomorphism density algorithms w.r.t. $K_3 \to G(n,\log^2 n/n)$}
\label{fig:running-time}
\end{figure}

We describe experiments which attempt to compare two currently available approaches to compute graph homomorphism features. The first method is the C++ implementation of the algorithm of D\'iaz et al.~\cite{diaz2002counting} given by the \texttt{homlib} library of NT and Maehara~\cite{ghc-nt20}. This algorithm includes a tree decomposition routine together with a dynamic programming algorithm to compute homomorphism numbers. Obtaining homomorphism densities is done by dividing by $n^k$ which is the cardinality of all morphisms from a pattern graph $F$ with $k$ nodes to a target graph $G$ with $n$ nodes. The second method is our implementation of Algorithm~\ref{alg:sample-hom-density} with two variants: exact edge membership queries to an adjacency list and approximate queries to a Bloom filter representing the edge set of $G$. Our implementation is written in the Rust programming language~\cite{matsakis2014rust} and will be made available as a library at a later date.
Our experiments use Erd\H{o}s-R\'enyi random graphs $G(n,p)$ to control the presence of subgraphs via the edge density $p$. This allows us to manipulate the target homomomorphism density for many small pattern graphs.  Furthermore, this method allows us to generate families of graphs of varying sizes which retain common properties. In Figure~\ref{fig:running-time} we report the running time of algorithms computing homomorphism densities from three small clique patterns: $K_3$, $K_4$, and $K_5$ to random graphs that are slightly above the connectivity threshold while remaining relatively sparse with an average degree of $\log^2 n$.

The \texttt{homlib} library implements graphs via adjacency lists and as such demonstrates great performance on smaller graphs. However, its running time is highly dependent on the homomorphism density of the target graph. %
Consider for example a random graph $G(1000,\num{9e-3})$ (disconnected w.h.p.) which contains very few triangles, for which \texttt{homlib} computes an exact triangle density of \num{7.9e-7} in only 8ms. The running time drastically increases to 102ms for $G(1000,\num{4e-2})$ (connected w.h.p.) which has a notably larger density of \num{6.4e-5}. More importantly, the running time of \texttt{homlib} scales linearly with the size of $G$ as shown in Figure~\ref{fig:running-time}. Surprisingly, increasing the treewidth parameter of the pattern graph $F$, from triangle to $K_5$ the complete graph over 5 nodes, incurs at most a linear increase in the running time which does not match the worst-case time complexity.

For the sample homomorphism density implementation of Algorithm~\ref{alg:sample-hom-density} we set a requested additive error of $\pm\num{1e-2}$ (this is not 1\% error) and $\pm\num{5e-3}$ with 95\% confidence, i.e. no more than 5\% of the sample densities exceed the error bound. In practice it is clear this theoretical error bound is extremely conservative. For example if we consider a graph $G(400,0.05)$ with approximately $4000$ edges, the exact triangle homomorphism density is \num{1.26e-4} which should be well below a ``detection threshold'' of $\pm\num{1e-2}$. However, the sample homomorphism density correctly identifies the order of magnitude with a sample density of \num{1.94e-4}.

In the case of our implementation of Algorithm~\ref{alg:sample-hom-density} with Bloom filters we select a 1\% false positive rate. Compared to the implementation with adjacency lists, the Bloom filter variant appears to be agnostic to the size of $G$. Surprisingly, we observe a behavior that is comparable to \texttt{homlib} in the sense that a low homomorphism density is associated with lesser running time. This can be explained by the fact that the less homomorphisms the higher the frequency of negative membership queries which short-circuits the entire group query. As noted in the analysis of its worst-case time complexity, the most important factor influencing the running time of Algorithm~\ref{alg:sample-hom-density} is the requested additive error $\varepsilon$. Figure~\ref{fig:running-time} reveals that a fixed additive error of $\varepsilon = \pm\num{1e-2}$ results in running times ranging from 30ms to 60ms while raising the precision to $\varepsilon = \pm\num{5e-3}$ increases the running time to values ranging from 100ms to 200ms. We underline that this additional cost is needed when the target homomorphism density becomes small. For example, the triangle homomorphism density of a random graph $G(n,\log^2 n)$ with $n = \num{e5}$ is \num{2.1e-8} which cannot be detected (the sample density is 0) with a precision of $\varepsilon = \pm\num{1e-2}$. However, a slightly better precision with $\varepsilon = \pm\num{5e-3}$ gives a coarse approximation of this value which is ``only'' off by 2 orders of magnitude. This detection threshold can also be used as an implicit filter discarding low-frequency patterns. Finally, we observe like with \texttt{homlib} that the size of $F$ has very little impact on the running time. %

We have conducted additional experiments with cuckoo filters~\cite{fan2014cuckoo} and XOR filters~\cite{graf2020xor} which reproduce the known tradeoffs between fast filter construction and fast filter querying. However the cuckoo filter implementation we have tested was slower approximately by 10\% to 20\% on both construction and querying than Bloom filters while XOR filters were up to 30\% faster than Bloom filters at the cost of more than double the construction time.

The above experimental results provide evidences that our implementation of Algorithm~\ref{alg:sample-hom-density} with Bloom filters would allow to scale a standard training pipeline including feature generation from existing datasets with graphs of \num{e2} nodes to currently unavailable datasets with graphs of \num{e5} nodes with no significant increase of the total running time, whereas using \texttt{homlib} would multiply the running time by at least 3 orders of magnitude.

\section{Conclusion}

In this paper, we have provided evidences that homomorphism numbers do not scale to large graphs due to limitations inherent to computational complexity but also from observations of the performance of practical implementations of state-of-art homomorphism counting algorithms. On the other hand we have shown that homomorphism numbers and homomorphism densities have equal representational power from a theoretical point of view. This is also true in practice when comparing homomorphism numbers and approximate homomorphism densities as features for graph classification where simple models using them as input attain comparable performance to popular graph neural network architectures. Finally we have demonstrated in synthetic experiments that our high-performance implementation of a simple algorithm for approximate homomorphism densities using Bloom filters is highly scalable.

As future works, we are interested in studying the families of pattern graphs used to compute homomorphism densities and would like to explore whether these families could be learned. A second direction would be to explore model explainability and the relationship between learned model weights and homomorphism information. Furthermore, graph homomorphisms are known to form an algebra~\cite{lovasz2012large}, where the product of homomorphism numbers of $G_1$ and $G_2$ corresponds to the homomorphism of a graph product between $G_1$ and $G_2$. This invites further investigation into whether polynomial kernels or other non-linear models implicitly capture unseen homomorphism information as well as studying its potential impact on classification performance. Finally, given the scalability of our algorithm to large graphs, we we would like to construct datasets for graph classification that contain sufficiently large graphs in order to fairly compare the performance of commonly used models on large-scale graph data.

\bibliographystyle{abbrv}
\bibliography{references}
\newpage
\appendix

\end{document}